\documentclass[11pt]{article}
\usepackage{coling2018}
\usepackage{times}
\usepackage{url}
\usepackage{latexsym}
\usepackage{graphicx}
\usepackage{fancyhdr}
\usepackage{bm}
\usepackage{paralist}
\usepackage{ctable}
\usepackage{multirow}
\usepackage{booktabs}
\usepackage{amsmath,amsbsy,amsthm,amssymb}

\title{Variational Attention for Sequence-to-Sequence Models}

\author{Hareesh Bahuleyan\thanks{\ \ The first two authors contributed equally.}\hspace{0.4em}\footnotemark[2]\quad  Lili Mou\footnotemark[1]\hspace{0.4em}\footnotemark[3] \quad Olga Vechtomova\footnotemark[2]\hspace{0.2em}\quad Pascal Poupart\footnotemark[2]\\
  \footnotemark[2]\hspace{0.2em} University of Waterloo, Canada \\
  {\tt \{hpallika, ovechtomova, ppoupart\}@uwaterloo.ca}\\ 
    \footnotemark[3]\hspace{0.2em} AdeptMind Research, Toronto, Canada\\ {\tt doublepower.mou@gmail.com}\\ }
\date{}
\newcommand{\btheta}{{\bm \theta}}
\newcommand{\bphi}{{\bm \phi}}

\newcommand{\KL}{{\operatorname{KL}}}
\newcommand{\RNN}{{\operatorname{RNN}}}

\newcommand{\n}{^{(n)}}
\newcommand{\src}{^\text{(src)}}
\newcommand{\tar}{^\text{(tar)}}
\newcommand{\barh}{\bar{\bm h}\src}
\begin{document}
\maketitle

\renewcommand{\headrulewidth}{0pt}
\renewcommand{\footskip}{30pt}
\cfoot{In \textit{ Proceedings of COLING 2018}.  Also accepted by \textit{TADGM Workshop@ICML 2018} for presentation. }
\thispagestyle{fancy}

\begin{abstract}
The variational encoder-decoder (VED) encodes source information as a set of random variables using a neural network, which in turn is decoded into target data using another neural network. In natural language processing, sequence-to-sequence (Seq2Seq) models typically serve as encoder-decoder networks. When combined with a traditional (deterministic) attention mechanism, the variational latent space may be bypassed by the attention model, and thus becomes ineffective. In this paper, we propose a variational attention mechanism for VED, where the attention vector is also modeled as Gaussian distributed random variables. Results on two experiments show that, without loss of quality, our proposed method alleviates the bypassing phenomenon as it increases the diversity of generated sentences.\footnote{Code is available at \url{https://github.com/HareeshBahuleyan/tf-var-attention}}
\end{abstract}

\section{Introduction}\blfootnote{This work is licensed under a Creative Commons Attribution 4.0 International License. License
details: \url{http://creativecommons.org/licenses/by/4.0}/}
\label{sec:intro}

The variational autoencoder (VAE), proposed by \newcite{kingma2013auto}, \textit{encodes} data to latent (random) variables, and then \textit{decodes} the latent variables to reconstruct the input data. Theoretically, it optimizes a variational lower bound of the log-likelihood of the data. Compared with traditional variational methods such as mean-field approximation~\cite{VI}, VAE leverages modern neural networks and hence is a more powerful density estimator. Compared with traditional autoencoders~\cite{AE}, which are \textit{deterministic}, VAE populates hidden representations to a region (instead of a single point), making it possible to generate diversified data from the vector space~\cite{Vseq2seq} or even control the generated samples~\cite{control}. 

In natural language processing (NLP), recurrent neural networks (RNNs) are typically used as both the encoder and decoder, known as a sequence-to-sequence (Seq2Seq) model. Although variational Seq2Seq models are much trickier to train in comparison to the image domain, \newcite{Vseq2seq} succeed in training a sequence-to-sequence VAE and generating sentences from a continuous latent space. Such an architecture can further be extended to a variational encoder-decoder (VED) to transform one sequence into another with the ``variational'' property~\cite{VHRED,zhou2017morphological}. 

When applying attention mechanisms~\cite{bahdanau2014neural} to variational Seq2Seq models, however, we find the generated sentences are of less variety, implying that the variational latent space is ineffective. The attention mechanism summarizes source information as an \textit{attention vector} by weighted sum, where the weights are a learned probabilistic distribution; then the attention vector is fed to the decoder.  Evidence shows that attention significantly improves Seq2Seq performance in translation~\cite{bahdanau2014neural}, summarization~\cite{summarization}, etc. In variational Seq2Seq, however, the attention mechanism unfortunately serves as a ``bypassing'' mechanism. In other words, the variational latent space does not need to learn much, as long as the attention mechanism itself is powerful enough to capture source information.

In this paper, we propose a variational attention mechanism to address this problem. We model the attention vector as random variables by imposing a probabilistic distribution. We follow traditional VAE and model the prior of the attention vector by a Gaussian distribution, for which we further propose two plausible priors, whose mean is either a zero vector or an average of source hidden states.

We evaluate our approach on two experiments: question generation and dialog systems. Experiments show that the proposed variational attention yields a higher diversity than variational Seq2Seq with deterministic attention, while retaining high quality of generated sentences. In this way, we make VED work properly with the powerful attention mechanism.

In summary, the main contributions of this paper are two-fold: (1) We discover a ``bypassing'' phenomenon in VED, which could make the learning of variational space ineffective. (2) We propose a variational attention mechanism that models the attention vector as random variables to alleviate the above problem. To the best of our knowledge, we are the first to address the attention mechanism in variational encoder-decoder neural networks. Our model is a general framework, which can be applied for various text generation tasks.

\section{Background and Motivation}

In this section, we introduce the variational autoencoder and the attention mechanism. We also present a pilot experiment motivating our variational attention model.

\subsection{Variational Autoencoder (VAE)}

A VAE encodes data $\bm Y$ (e.g., a sentence) as hidden random variables $\bm Z$, based on which the decoder reconstructs $\bm Y$. 
Consider a generative model, parameterized by $\bm \theta$, as 
\begin{equation}
p_{\bm\theta}(\bm Z,\bm Y)=p_{\bm{\theta}}(\bm Z)p_{\bm\theta}(\bm Y|\bm Z)
\end{equation}

Given a dataset $\mathcal{D}=\{\bm y\n\}_{n=1}^N$, the likelihood of a data point is
\begin{align}
  &\log p_\btheta(\bm y\n) \ge \begin{aligned}[t]
      &\mathbb{E}_{\bm z\sim q_\bphi(\bm z|\bm y\n)}\left[\log\left\{\dfrac
{p_\btheta(\bm y\n, \bm z)}
{q_\bphi(\bm z|\bm y\n)}
\right\}\right]\\\nonumber
       \end{aligned}\\
  &= \begin{aligned}[t]
      &\mathbb{E}_{\bm z\sim q_\bphi(\bm z|\bm y\n)}\left[
\log p_\btheta(\bm y\n|\bm z)
\right]-\KL\left(q_\bphi(\bm z|\bm y\n)\|p(\bm z)\right)\overset{\Delta}{=}\mathcal{L}\n(\btheta,\bphi)
       \end{aligned}\label{eqn:VAElb}
\end{align}

VAE models both $q_\bphi(\bm z|\bm y)$ and $p_\btheta(\bm y|\bm z)$ with neural networks, parametrized by $\bphi$ and $\btheta$, respectively. Figure~\ref{fig:GM}a shows the graphical model of this process. The training objective is to maximize the lower bound of the likelihood $\mathcal{L}(\btheta,\bphi)$, which can be rewritten as minimizing
\begin{align}
J\n = J_\text{rec}(\btheta,\bphi,\bm y\n) + \KL\!\left(\!q_\bphi(\bm z|\bm y\n)\|p(\bm z)\!\right)\label{eqn:loss}
\end{align}
The first term, called \textit{reconstruction loss}, is the (expected) negative log-likelihood of data, similar to traditional deterministic autoencoders. The expectation is obtained by Monte Carlo sampling. The second term is the KL-divergence between $\bm z$'s posterior and prior distributions. Typically the prior is set to standard normal $\mathcal{N}(\bm 0, \mathbf {I})$.

\subsection{Variational Encoder-Decoder (VED)}
In some applications, we would like to transform source information to target information, e.g., machine translation, dialogue systems, and text summarization. In these tasks, ``auto"-encoding is not sufficient, and an encoding-decoding framework is required.
Different efforts have been made to extend VAE to variational encoder-decoder (VED) frameworks, which transform an input $\bm X$ to output $\bm Y$. One possible extension is to condition all probabilistic distributions further on $\bm X$~\cite{zhang2016variational,cao2017latent,VHRED}. In this case, the posterior of $\bm z$ is given by $q_\bphi(\bm z|\bm X, \bm Y)$. This, however, introduces a discrepancy between training and prediction, since $\bm Y$ is not available during the prediction stage.

\begin{figure}[!t]
	\centering
	\includegraphics[width=0.5\linewidth]{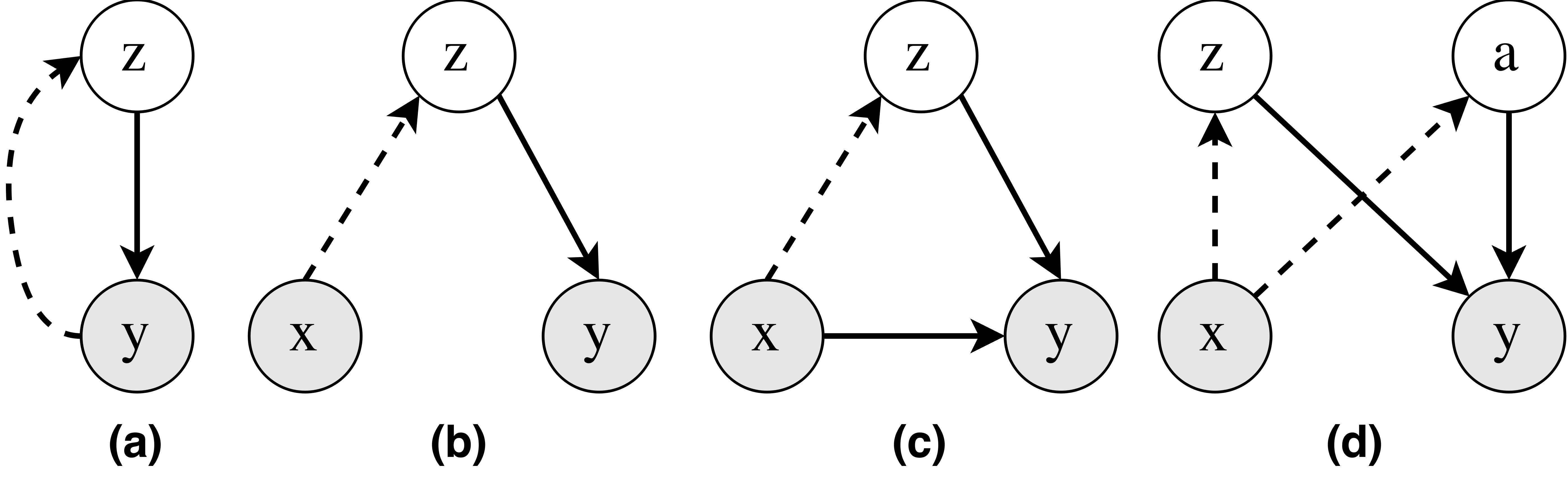}
	\caption{Graphical model representations. \textbf{(a)} Variational autoencoder (VAE). \textbf{(b)} Variational encoder-decoder (VED). \textbf{(c)} VED with deterministic attention (VED+DAttn). \textbf{(d)} VED with variational attention (VED+VAttn). \textbf{Dashed lines}: Encoding phase. \textbf{Solid lines}: Decoding phase.}
	\label{fig:GM}
\end{figure}

Another approach is to build a recognition model using only $\bm X$ \cite{zhou2017morphological}.  Making the assumption that $\bm Y$ is a function of $\bm X$, i.e., $\bm Y=\bm Y(\bm X)$, we have $q_\bphi(\bm z|\bm y)=q_\bphi(\bm z|\bm Y(\bm x))\overset{\Delta}{=}q_\bphi(\bm z|\bm x)$. In this work, we follow \newcite{zhou2017morphological} and adopt this extension. Figure~\ref{fig:GM}b shows the graphical model of the VED used in our work.

\subsection{Seq2Seq and Attention Mechanism}

In NLP, sequence-to-sequence recurrent neural networks are typically used as the encoder and decoder, as they are suitable for modeling a sequence of words (i.e., sentence). Figure~\ref{fig:models}a shows a basic Seq2Seq model in the VAE/VED scenario~\cite{Vseq2seq}. The encoder has an input $\bm x$, and outputs $\bm \mu_z$ and $\bm \sigma_z$ as the parameters of $\bm z$'s posterior normal distribution. Then a decoder generates $\bm y$ based on a sample $\bm z$, drawn from its posterior distribution.
 
Attention mechanisms are proposed to dynamically align $\bm y=(y_1, \cdots, y_{|\bm y|})$ and $\bm x=(x_1,\cdots, x_{|\bm x|})$ during generation. At each time step $j$ in the decoder, the attention mechanism computes a probabilistic distribution by 
\begin{equation}
\alpha_{ji}=\frac{\exp\{\widetilde{\alpha}_{ji}\}}{\sum_{i'=1}^{|\bm x|}\exp\{\widetilde{\alpha}_{ji'}\}}
\end{equation}
where $\widetilde{\alpha}_{ji}$ is a pre-normalized score, computed by $\widetilde{\alpha}_{ji}=\bm h\tar_j W^T\bm h\src_i$ in our model. Here, $\bm h\tar_j$ and $\bm h\src_i$ are the hidden representations of the $j$th step in target and $i$th in the source, and $W$ is a learnable weight matrix. 

Then the source information $\{\bm h_i\src\}_{i=1}^{|\bm x|}$ is summed by weights $\alpha_{ji}$ to obtain the attention vector
\begin{equation}
\bm a_j = \sum_{i=1}^{|\bm x|} \alpha_{ji}\bm h\src_i\label{eqn:att}
\end{equation}
which is fed to the decoder RNN at the $j$th step. Figure~\ref{fig:models}b shows the variational Seq2Seq model with such traditional attention.

\begin{table}[!t]
  \centering
  \resizebox{0.95\linewidth}{!}{
    \begin{tabular}{l|l}
    \toprule
    \multicolumn{2}{c}{\textbf{Input}: \textit{the men are playing musical instruments}} \\
    \midrule
    \textbf{(a) VAE w/o hidden state init. (Avg entropy: 2.52)} & \textbf{(b) VAE w/ hidden state init. (Avg entropy: 2.01)} \\
    \midrule
    \textit{the men are playing musical instruments} & \textit{the men are playing musical instruments} \\
    \textit{the men are playing video games} & \textit{the men are playing musical instruments} \\
    \textit{the musicians are playing musical instruments} & \textit{the men are playing musical instruments} \\
    \textit{the women are playing musical instruments} & \textit{the man is playing musical instruments} \\
    \bottomrule
    \end{tabular}%
  }
  \label{tab:addlabel}%
  \caption{Sentences obtained by sampling from the VAE's latent space. (a) VAE without hidden state initialization. (b) VAE with hidden state initialization.}
  \label{tab:motivation}%
\end{table}%

\subsection{``Bypassing'' Phenomenon}\label{ss:motivation}

In this part, we explain the ``bypassing'' phenomenon in VAE/VED, if the network is not designed properly; this motivates our variational attention described in Section~\ref{sec:model}.

We observe that, if the decoder has a direct, deterministic access to the source, the latent variables $\bm Z$ might not capture much information so that the VAE or VED does not play a role in the process. We call this a \textit{bypassing phenomenon}.

Theoretically, if $p_\btheta(\bm Y|\cdot)$ is aware of $\bm X$ by itself, i.e.,  $p_\btheta(\bm Y|\cdot)$ becomes $p_\btheta(\bm Y|\bm X, \bm Z)$, it could be learned as $p_\btheta(\bm Y|\bm X)$ without hurting the reconstruction loss $J_\text{rec}$, but the $\KL$ term in Eq.~(\ref{eqn:loss}) can be minimized by fitting the posterior to its prior. This degrades a variational Seq2Seq model to a deterministic one.

\begin{figure}[!t]
	\centering
	\includegraphics[width=.85\linewidth]{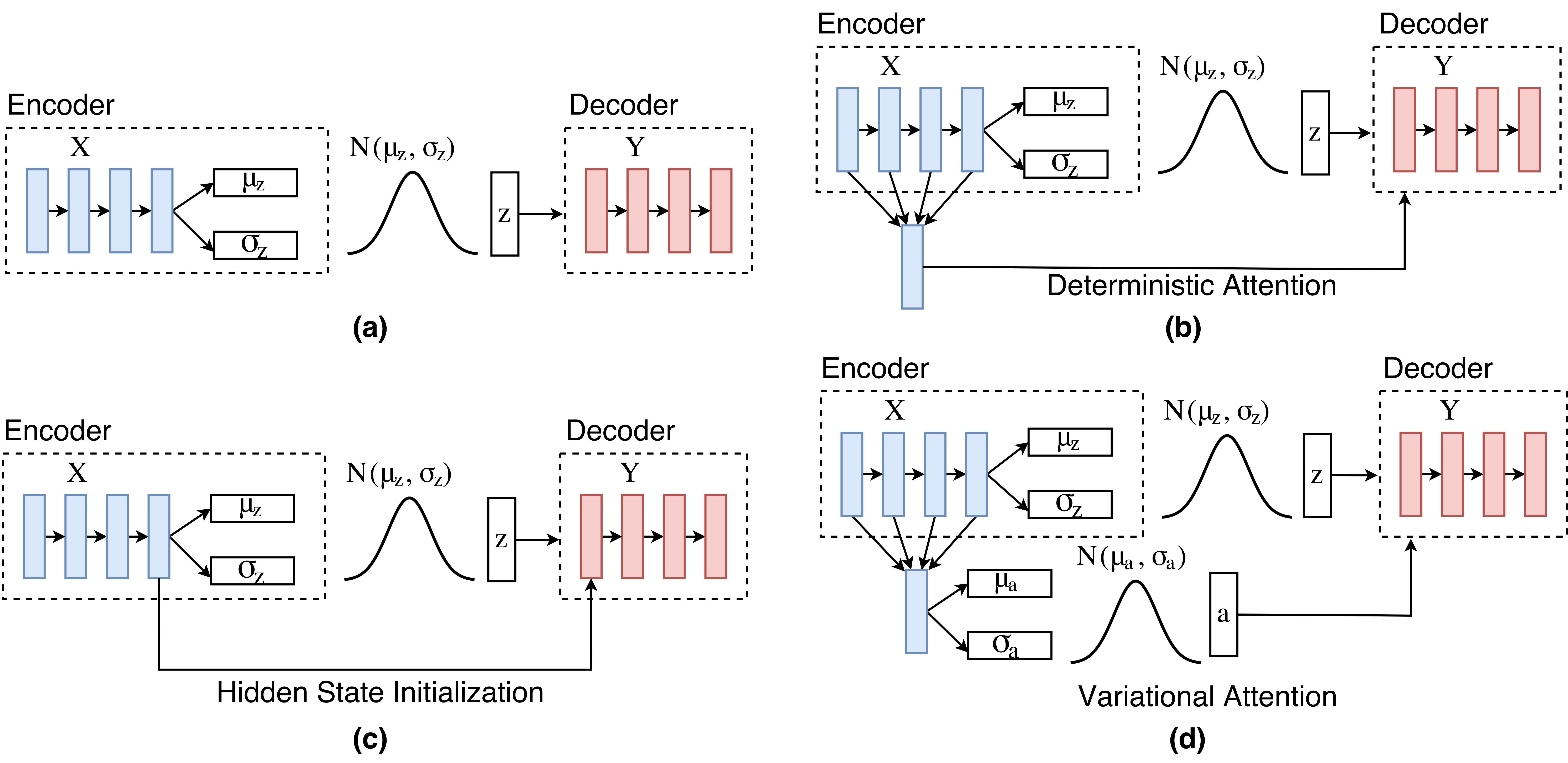}
	\caption{(a) Variational Seq2Seq model. (b) Variational Seq2Seq with deterministic attention. (c) Variational Seq2Seq with hidden state initialization. (d) Variational Seq2Seq with variational attention.}\label{fig:models}
\end{figure}

The phenomenon can be best shown with a bypassing connection between the encoder and decoder for hidden state initialization. Some previous studies using VEDs set the decoder's initial state to be the encoder's final state~\cite{cao2017latent}, shown in Figure~\ref{fig:models}c. We conducted a pilot study with a Seq2Seq VAE with a subset ($\sim$80k samples) of the massive dataset provided by~\newcite{SNLI}, and show generated sentences and entropy in Table~\ref{tab:motivation}. We see that the variational Seq2Seq can only generate very similar sentences with such bypassing connections (Table~\ref{tab:motivation}b), as opposed to generating diversified samples from the latent space (Table~\ref{tab:motivation}a). We also computed the entropy for 10 randomly sampled outputs for a given input sentence. Quantitatively, the entropy decreases by 0.5 on average for 1k unseen input sentences. This shows a significant difference because entropy is a logarithmic metric. Our analysis sheds light on the design philosophy of neural architectures in VAE or VED. 
 
Since attention largely improves model performance for deterministic Seq2Seq models, it is tempting to include attention in the variational Seq2Seq as well. However, our pilot experiment raises the doubt if a traditional attention mechanism, which is deterministic, may bypass the latent space in VED, as illustrated by a graphical model in Figure~\ref{fig:GM}c. Also, evidence in \newcite{pastfuture} shows the attention mechanism is so powerful that removing other connections between the encoder and decoder has little effect on BLEU scores in machine translation. Therefore, a VED with deterministic attention might learn reconstruction mostly from attention, whereas the posterior of the latent space can fit to its prior in order to minimize the $\KL$ term.

To alleviate this problem, we propose a variational attention mechanism for variational Seq2Seq models, as is described in detail in the next section.

\section{The Proposed Variational Attention}
\label{sec:model}

Let us consider the decoding process of an RNN. At each timestep $j$, it adjusts its hidden state $\bm h\tar_j$ with an input of a word embedding $\bm y_{j-1}$ (typically the groundtruth during training and the prediction from the previous step during testing). This is given by $\bm h\tar_j = \RNN_\btheta(\bm h\tar_{j-1}, \bm y_{j-1})$. In our experiments, we use long short-term memory units~\cite{hochreiter1997long} as RNN's transition. Enhanced with attention, the RNN is computed by $\bm h\tar_j = \RNN_\btheta(\bm h\tar_{j-1}, [\bm y_{j-1}; \bm a_j])$. The predicted word is given by a softmax layer $p(y_j)=\operatorname{softmax}(W_\text{out}\bm h\tar_j)$, where $W_\text{out}$ is a weight matrix. As discussed earlier, traditional attention computes $\bm a_j$ in a deterministic fashion by Eq.~(\ref{eqn:att}).

To build a variational attention, we treat both the traditional latent space $\bm z$ and the attention vector $\bm a_j$ as random variables. The recognition and reconstruction graphical models are shown in Figure~\ref{fig:GM}d.

\subsection{Lower Bound}

Since the likelihood of the $n$th data point decomposes for different time steps, we consider the lower bound $\mathcal{L}_j\n(\btheta, \bphi)$ at the $j$th step. The variational lower bound in Eq.~(\ref{eqn:VAElb}) becomes
\begin{align}
  \mathcal{L}_j\n(\btheta, \bphi) &= \begin{aligned}[t]
      &\mathbb{E}_{\bm z,\bm  a\sim q_\bphi(\bm z, \bm a|\bm x\n)}\left[
\log p_\btheta(\bm y\n|\bm z, \bm a)
\right]-\KL\left(q_\bphi(\bm z, \bm a|\bm x\n)\|p(\bm z, \bm a)\right)
       \end{aligned}\\
  &= \begin{aligned}[t]
      &\mathbb{E}_{\bm z\sim q^{(z)}_\bphi(\bm z|\bm x\n), \bm a\sim q^{(a)}_\bphi(\bm a|\bm x\n)}\left[
\log p_\btheta(\bm y\n|\bm z, \bm a)
\right]\\
&-\KL\left(q^{(z)}_\bphi(\bm z|\bm x\n)\|p(\bm z)\right)-\KL\left(q^{(a)}_\bphi(\bm a|\bm x\n)\|p(\bm a)\right)
       \end{aligned}\label{eqn:lb2}
\end{align}
Eq.~(\ref{eqn:lb2}) is due to the independence in both recognition and reconstruction phrases. The posterior factorizes as $q_\phi(\bm z, \bm a|\cdot)=q_\phi^{(z)}(\bm z|\cdot)\ q_\phi^{(a)}(\bm a|\cdot)$ because $\bm z$ and $\bm a$ are conditionally independent given $\bm x$ (dashed lines in Figure~\ref{fig:GM}d), whereas the prior factorizes because $\bm z$ and $\bm a$ are marginally independent (solid lines in Figure~\ref{fig:GM}d). In this way, the sampling procedure can be done separately and the $\KL$ loss can also be computed independently.

\subsection{Prior}
We propose two plausible prior distributions for $\bm a_j$.
\begin{compactitem}
	\item The simplest prior, perhaps, is the standard normal, i.e., $p(\bm a_j)=\mathcal{N}(\bm 0, \mathbf{I})$. This follows the prior of the latent space $\bm z$ as in a conventional autoencoder~\cite{kingma2013auto,Vseq2seq}.
	\item We observe that the attention vector has to be inside the convex hull of hidden representations of the source sequence, i.e., $\bm a_j\in\operatorname{conv}\{\bm h\src_i\}$. We impose a normal prior whose mean is the average of $\bm h\src_i$, i.e., $p(\bm a_j)=\mathcal{N}(\bar{\bm h}\src, \mathbf{I})$, where $\bar{\bm h}\src = \frac{1}{|\bm x|}\sum_{i=1}^{|\bm x|}\bm h_i\src$, making the prior non-informative.
\end{compactitem}
	
\subsection{Posterior}
We model the posterior of $q_\bphi^{(a)}(\bm a_j|\bm x)$ as a normal distribution $\mathcal{N}(\bm \mu_{a_j}, \bm\sigma_{a_j})$, where the parameters $\bm \mu_{a_j}$ and $\bm\sigma_{a_j}$ are obtained by a recognition neural network. Similar to VAEs, we compute parameters as if in the deterministic attention in Eq.~(\ref{eqn:att}) (denoted by $\bm a_j^\text{det}$ in this part)  and then transform them by another layer, shown in Figure~\ref{fig:models}d. 

For the mean $\bm \mu_{a_j}$, we apply an identity transformation, i.e., $\bm \mu_{a_j}\equiv \bm a_j^\text{det}$. The identify transformation makes much sense as it preserves the spirit of ``attention.'' To compute $\bm \sigma_{a_j}$, we first transform $\bm a_j^\text{det}$ by a neural layer with $\tanh$ activation. The resulting vector then undergoes a linear transformation followed by an $\exp$ activation function to ensure that the values are positive.

\subsection{Training Objective}

The overall training objective of Seq2Seq with both variational latent space $\bm z$  and variational attention $\bm a$ is to minimize
\begin{align}
J\n(\btheta,\bphi)&=J_\text{rec}(\btheta,\bphi, \bm y\n)+ \lambda_\KL\Big[\KL\left(q_\phi^{(z)}(\bm z|\bm x\n)\|p(\bm z)\right)+\gamma_a\sum_{j=1}^{|\bm y|}
\KL\left(q_\phi^{(a)}(\bm a_j|\bm x\n)\|p(\bm a_j)
\right)
\Big]\label{eqn:vatt.obj}
\end{align}
 
Here, we have a hyperparameter $\lambda_\KL$ to balance the reconstruction loss and KL losses. $\gamma_a$ further balances the attention's KL loss and $\bm z$'s KL loss. Since VAE and VED are tricky with Seq2Seq models (e.g., requiring KL annealing), we tie the change of both KL terms and only anneal $\lambda_\KL$. (Training details will be presented in Section~\ref{sec:details}.)

Notice that if $\bm a_j$ has a prior of $\mathcal{N}(\barh, \mathbf{I})$, the derivative of the KL term also goes to $\barh$. This can be computed straightforwardly or by auto-differentiation tools, e.g., TensorFlow.

\subsection{Geometric Interpretation}
\begin{figure*}
\centering
	\includegraphics[width=.8\textwidth]{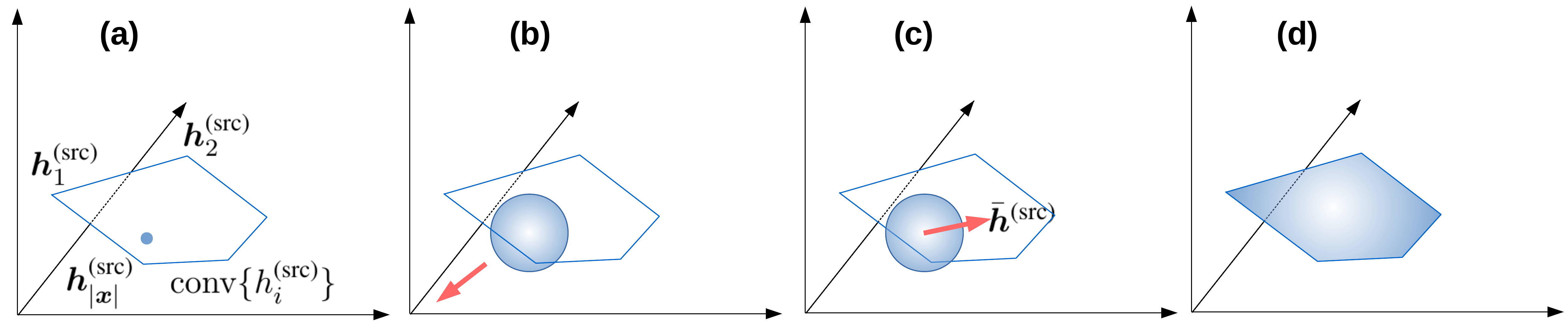}
	\caption{Geometric interpretation of attention mechanisms.}\label{fig:geo}
\end{figure*}

We present a geometric interpretation of both deterministic and variational attention mechanisms in Figure~\ref{fig:geo}.

Suppose the hidden representations $\bm h_i\src$ is of $k$-dimensional space (represented as a 3-d space in Figure~\ref{fig:geo}). In the deterministic mechanism, the attention model is a convex combination of $\{\bm h_i\src\}_{i=1}^{|\bm x|}$, as the weights in Eq.~(\ref{eqn:att}) are a probabilistic distribution. The attention vector $\bm a_j$ is a point in the convex hull $\operatorname{conv}\{\bm h_i\src\}$, shown in Figure~\ref{fig:geo}a.

For variational attention in Figures~\ref{fig:geo}b and~\ref{fig:geo}c, the mean of posterior is still in the convex hull, but the sample drawn from the posterior is populated over the entire space (although mostly around the mean, shown as a ball). The difference between the two variants is that the standard normal prior $\mathcal{N}(\bm 0, \mathbf{I})$ pulls the posterior to the origin, whereas the prior $\mathcal{N}(\barh, \mathbf{I})$ pulls the posterior to the mean of $\bm h_1\src,\bm h_2\src, \cdots, \bm h_{|\bm x|}\src$ (indicated by red arrows).

Finally we would like to present a (potential) alternative of modeling variational attention. Instead of treating $\bm a_j$ as random variables, we might also treat $\bm \alpha_j$ as random variables. Since $\bm \alpha_j$ is the parameter of a categorical distribution, its conjugate prior is a Dirichlet distribution. In this case, the resulting attention vector populates the entire convex hull (Figure~\ref{fig:geo}d). However, it relies on a reparametrization trick to propagate reconstruction error's gradient back to the recognition neural network~\cite{kingma2013auto}. In other words, the sampling of latent variables should be drawn from a fixed distribution (without parameters) and then transformed to a desired sample using the distribution's parameters. This is nontrivial for Dirichlet distributions and further research is needed to address this problem.

\section{Experiments}

We evaluated our model on two tasks: question generation (Section~\ref{ss:qg}) and dialog systems (Section~\ref{ss:dialog}).

\subsection{Experiment I: Question Generation}\label{ss:qg}
\begin{table*}[!t]
  \centering
  \resizebox{.97\textwidth}{!}{
    \begin{tabular}{l|c|ccccccc}
    \toprule
    \textbf{Model} & \textbf{Inference} & \textbf{BLEU-1} & \textbf{BLEU-2} & \textbf{BLEU-3} & \textbf{BLEU-4} & \textbf{Entropy} & \textbf{Dist-1} & \textbf{Dist-2} \\
    \midrule
    DED (w/o Attn) \cite{Qgen} & MAP   & 31.34 & 13.79 & 7.36 & 4.26 & -     & -     & - \\
    \midrule
    DED (w/o Attn)& MAP   & 29.31 & 12.42 & 6.55 & 3.61  & -     & -     & - \\
    DED+DAttn & MAP   & 30.24 & 14.33 & 8.26 & 4.96 & -     & -     & - \\
    \midrule
    \multirow{2}{*}{VED+DAttn} 	& MAP   & \textbf{31.02} & 14.57 & 8.49 & 5.02 & -  & -  & - \\
						          & Sampling & 30.87 & \textbf{14.71} & \textbf{8.61} & \textbf{5.08} & 2.214  & 0.132  & 0.176 \\
	\midrule
    \multirow{2}{*}{VED+DAttn (2-stage training)} 	& MAP   & 28.88 & 13.02 & 7.33 & 4.16 & -  & -  & - \\
						        & Sampling & 29.25 & 13.21 & 7.45 & 4.25 & 2.241  & 0.140  & 0.188 \\
    \midrule
    \multirow{2}{*}{VED+VAttn-$0$} & MAP   & 29.70 & 14.17 & 8.21 & 4.92 & -  & -  & - \\
                                   & Sampling & 30.22 & 14.22 & 8.28 & 4.87 & \textbf{2.320}  & \textbf{0.165}  & \textbf{0.231} \\
    \midrule
    \multirow{2}{*}{VED+VAttn-$\bar{h}$} & MAP   & 30.23 & 14.30 & 8.28 & 4.93 & -  & -  & - \\
                                  & Sampling & 30.47 & 14.35 & 8.39 & 4.96 & 2.316 & 0.162 & 0.228 \\
    \bottomrule
    \end{tabular}
}
    \caption{BLEU, entropy, and distinct scores. We compare the deterministic encoder-decoder (DED) and variational encoder-decoders (VEDs). For VED, we have several variates: deterministic attention (DAttn) and the proposed variational attention (VAttn). Variational models are evaluated by both max \textit{a posteriori} (MAP) inference and sampling.}
  \label{tab:result}%
\end{table*}

\paragraph{Task, Dataset, and Metrics.} We first evaluated our approach on a question generation task. It uses the Stanford Question Answering Dataset ~\cite[SQuAD]{squad}, and aims to generate questions based on a sentence in a paragraph. 
We used the same train-validation-test split as in \newcite{Qgen}. According to \newcite{Qgen}, the attention mechanism is especially critical in this task in order to generate relevant questions. Also, generated questions do need some variety (e.g., in the creation of reading comprehension datasets), as opposed to machine translation, which is typically deterministic.

We followed \newcite{Qgen} and used BLEU-1 to BLEU-4 scores \cite{papineni2002bleu} to evaluate the quality (in the sense of accuracy) of generated sentences. Besides, we adopted entropy and \textit{distinct} metrics to measure the diversity. Entropy is computed as $-\sum_w p(w)\log p(w)$, where $p(\cdot)$ is the unigram probability in generated sentences. \textit{Distinct} metrics---used in previous works to measure diversity~\cite{distinct}---computes the percentage of distinct unigrams or bigrams (denoted as \textit{Dist-1} and \textit{Dist-2}, respectively).

\paragraph{Training Details.}
We used LSTM-RNNs with 100 hidden units for both the encoder and decoder; the dimension of the latent vector $\bm z$ was also 100d. We adopted 300d word embeddings \cite{word2vec}, pretrained on the SQuAD dataset. For both the source and target sides, the vocabulary was limited to the most frequent 40k tokens. We used the Adam optimizer \cite{adam} to train all models, with an initial learning rate of 0.005, a multiplicative decay of 0.95, and other default hyperparameters. The batch size was set to be 100.

As shown in \newcite{Vseq2seq}, Seq2Seq VAE is hard to train because of the issues associated with the $\KL$ term vanishing to zero. Following \newcite{Vseq2seq}, we adopted $\KL$ cost annealing and word dropout during training. The coefficient of the KL term $\lambda_{\KL}$ was gradually increased using a logistic annealing schedule, allowing the model to learn to reconstruct the input accurately during the early stages of training. A fixed word dropout rate of $25\%$ was used. 

All the hyperparameter tuning was based on validation performance on the motivating Seq2Seq VAE discussed in Section~\ref{ss:motivation}, and the same hyperparameters were used for all of the models described in Section~\ref{sec:model}. 

\label{sec:details}

\paragraph{Overall Performance.}

Table~\ref{tab:result} represents the performance of various models. We first implemented a traditional vanilla Seq2Seq model, which we call a deterministic encoder-decoder (DED), and generally replicated the results on the question generation task as reported in \newcite{Qgen}, showing that our implementation is fair. 
Incorporating attention mechanism in this model (DED+DAttn) improves BLEU scores, as expected.
In the variational encoder-decoder (VED) framework, we report results obtained by both max \textit{a posterior} (MAP) inference as well as sampling. In the sampling setting, we draw 10 samples ($\bm z$ and/or $\bm a$) from the posterior given $\bm x$ for each data point, and report average BLEU scores.

The proposed variational attention model (VED+VAttn) largely outperforms deterministic attention (VED+DAttn) in terms of all diversity metrics. It should be noted that entropy is a logarithmic measure, and hence a difference of 0.1 in Table~\ref{tab:result} is significant; VED+VAttn also generates more distinct unigrams and bigrams than VED+DAttn.

Regarding the prior of variational attention, we propose two variants: $\mathcal{N}(\bm 0, \mathbf{I})$ and $\mathcal{N}(\barh, \mathbf{I})$, denoted as VED+VAttn-0 and VED+VAttn-$\bar h$, respectively. VED+VAttn-0 has slightly lower BLEU but higher diversity. The results are generally comparable, showing both priors are reasonable.

We also tried a heuristic of 2-stage training (VED+DAttn 2-stage), in which the VED is first trained without attention for 6 epochs, and then the attention mechanism is added to the model. This heuristic is proposed in hopes of better training the variational latent space at the beginning stages. However, experiments show that such simple heuristic does not help much, and is worse than the principled variational attention mechanism in terms of all BLEU and diversity metrics.

\begin{figure*}[t]
\centering
\includegraphics[width=.9\textwidth]{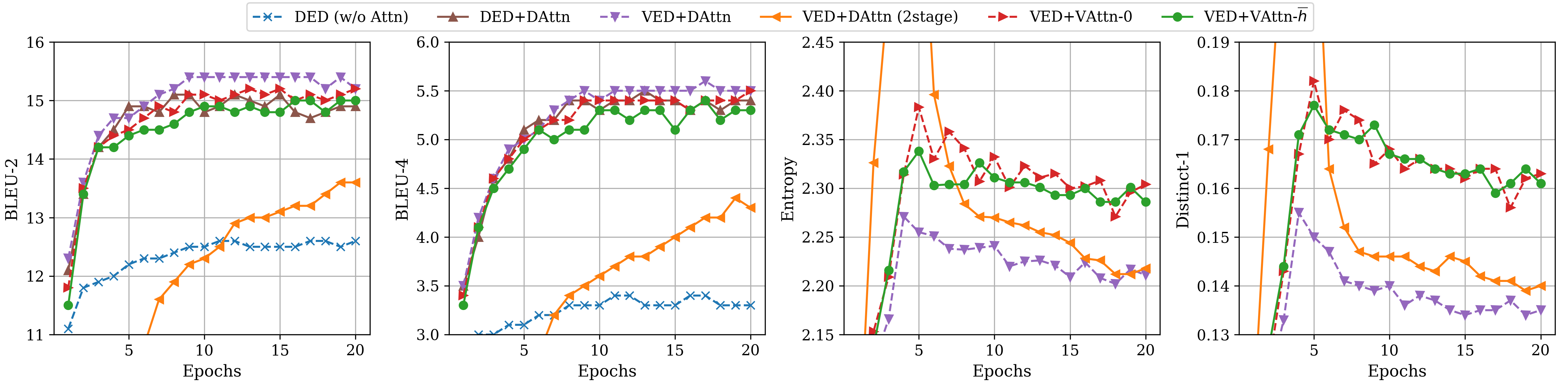}
\vspace{-.5em}
	\caption{BLEU-2, BLEU-4, Entropy, and \textit{Dist-1} calculated on the validation set as training progresses.}\label{fig:model-trends}
\end{figure*}
\begin{figure*}[t]
\centering
\includegraphics[width=.9\textwidth]{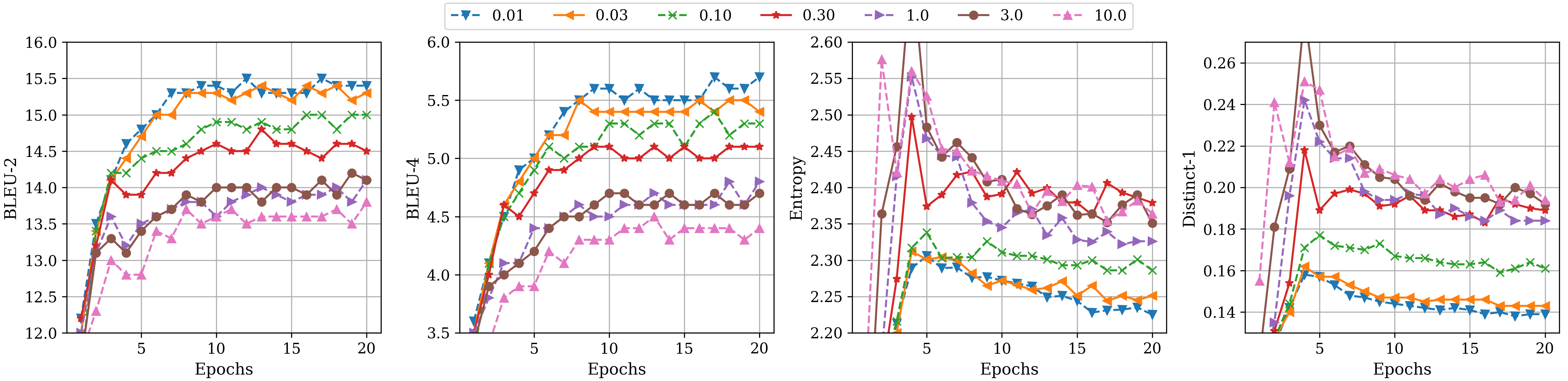}
\vspace{-.5em}
	\caption{BLEU-2, BLEU-4, Entropy, and \textit{Dist-1} with different $\gamma_a$ values.}\label{fig:gamma-sensitivity}
\end{figure*}

\paragraph{Human Evaluation.} In order to assess the quality of the generated text in terms of language fluency, a human evaluation study was carried out. 
For each of the two models under comparison (VED+DAttn and VED+VAttn-$\bar h$), a randomly shuffled subset of 100 generated questions were selected. Six human evaluators were asked to rate the fluency of these 200 questions on a 5-point scale: 5-Flawless, 4-Good, 3-Adequate, 2-Poor, 1-Incomprehensible, following \newcite{stent2005evaluating}. The average rating obtained for VED+DAttn was 3.99 and for VED+VAttn-$\bar h$ was 4.01, the difference between which is not statistically significant. The human annotations achieved 0.61 average Spearman correlation coefficient (measuring order correlation) between any two annotators. According to \newcite{swinscow1976statistics}, this indicates {moderate to strong} correlation among different annotators. Hence, we conclude variational attention does not negatively affect the fluency of sentences.

\paragraph{Learning curves.} 
Figure~\ref{fig:model-trends} shows the trends of sentence quality (BLEU-2 and BLEU-4) and diversity (entropy and \textit{Dist-1}) of all models on the validation set, as training progresses.\footnote{Other metrics are omitted because the trend is the same.} 
We see that BLEU and diversity are conflicting objectives: a high BLEU score indicates resemblance to the groundtruth, resulting in low diversity. However, the variational attention mechanisms (red and green lines in Figure~\ref{fig:model-trends}) remain high in both aspects, showing the effectiveness of our model.

\begin{table}[!t]
	\centering
    \resizebox{\textwidth}{!}{
	\begin{tabular}{|rl|}
		\toprule
		\textbf{Source}\!\!\!& \textit{when the british forces evacuated at the close of the war in 1783 ,\!\!\!} \\
		&they transported 3,000 freedmen for resettlement in nova scotia .\\
		\textbf{Reference}\!\!\!& \textit{in what year did the american revolutionary war end ?} \\
		\midrule
		\multirow{3}[2]{*}{\textbf{VED+DAttn}}\!\!\!& \textit{how many people evacuated in newfoundland ?} \\
		& \textit{how many people evacuated in newfoundland ?} \\
		& \textit{what did the british forces seize in the war ?} \\
		\midrule
		\!\!\!\multirow{3}[2]{*}{\textbf{VED+Vattn}-$\bar{\bm h}$}\!\!\!& \textit{how many people lived in nova scotia ?} \\
		& \textit{where did the british forces retreat ?} \\
		& \textit{when did the british forces leave the war ?} \\
		\bottomrule
    \end{tabular}
	\begin{tabular}{|rl|}
		\toprule
		\textbf{Source}\!\!\!& \textit{downstream , more than 200,000 people were evacuated from} \\
		& \textit{mianyang by june 1 in anticipation of the dam bursting .}\\
		\textbf{Reference}\!\!\!& \textit{how many people were evacuated downstream ?} \\
		\midrule
		\multirow{3}[2]{*}{\textbf{VED+DAttn}}\!\!\!& \textit{how many people evacuated from the mianyang basin ?} \\
		& \textit{how many people evacuated from the mianyang basin ?} \\
		& \textit{how many people evacuated from the mianyang basin ?} \\
		\midrule
		\!\!\!\multirow{3}[2]{*}{\textbf{VED+VAttn}-$\bar{\bm h}$}\!\!\!& \textit{how many people evacuated from the tunnel ?} \\
		& \textit{how many people evacuated from the dam ?} \\
		& \textit{how many people were evacuated from fort in the dam ?} \\
		\bottomrule
	\end{tabular}
}
	\caption{Case study of question generation.}
	\label{tab:case}%
\end{table}%
\paragraph{Strength of Attention's KL Loss.}
We tuned the KL term's strength in variational attention, i.e., $\gamma_a$ in Eq.~(\ref{eqn:vatt.obj}), and plot the BLEU and diversity metrics in Figure~\ref{fig:gamma-sensitivity}. In this experiment, we used the VED+DAttn-$\bar h$ variant. As shown, a decrease in $\gamma_a$ increases the quality of generated sentences at the cost of diversity. This is expected because a lower $\gamma_a$ gives the model less incentive to optimize the attention's KL term, which then causes the model to behave more ``deterministic.'' Based on this experiment, we chose a value of 0.1 for $\gamma_a$, as it yields a learning curve in the middle among those of different hyperparameters, being a good balance between quality and diversity.

It should be further mentioned that, with a milder $\gamma_a$ (e.g., 0.01), VED+VAttn outperforms VED+DAttn in terms of both quality and diversity (on the validation set). This is consistent with the evidence that variational latent space may serve as a way of regularization and improves quality~\cite{zhang2016variational}. However, a small $\gamma_a$ only slightly improves diversity, and hence we did not choose this hyperparameter in Table~\ref{tab:result}.

\paragraph{Case study.} We show in Table~\ref{tab:case} two examples of generated sentences by VED+DAttn and VED+VAttn-$\bar h$, each containing three random sentences drawn from the variational latent space(s) for a given input. In both examples, the variational attention generates more diversified sentences than deterministic attention. The quality of generated sentences is close in both models.

\subsection{Experiment II: Dialog Systems}\label{ss:dialog}

\begin{table}[!t]
	\centering
	\resizebox{.5\textwidth}{!}{
	\begin{tabular}{l|c|cccc}
		\toprule
		\textbf{Model} & \textbf{Inference}& \textbf{BLEU-2} & \textbf{Entropy} & \textbf{Dist-1} & \textbf{Dist-2}\\
		\midrule
		DED+DAttn & MAP & 1.84 & -- & -- & --\\\midrule
		\multirow{2}{*}{VED+DAttn} & MAP & 1.68& -- & -- & --\\
		& Sampling & 1.68 & 2.113 & 0.311 & 0.450\\\midrule
		\multirow{2}{*}{VED+VAttn-$\bar h$}& MAP & 1.78 & -- & -- & --\\
		& Sampling& 1.79 & \textbf{2.167} & \textbf{0.324} & \textbf{0.467}\\
		\bottomrule
	\end{tabular}
}
\caption{Performance on conversation systems.}\label{tab:result2}
\end{table}
We present another experiment on generative conversation systems. The goal is to generate a reply based on a user-issued utterance. We used the Cornell Movie-Dialogs Corpus\footnote{\url{https://www.cs.cornell.edu/~cristian/Cornell_Movie-Dialogs_Corpus.html}}~\cite{cornell} as our dataset, which contains more than 200k conversational exchanges. All the settings in this experiment were the same as in Subsection~\ref{ss:qg} except that we had 30k words as the vocabulary for both the encoder and decoder. We evaluated the  quality of generated replies with BLEU-2, as it has been observed to be more or less correlated with human annotators among the BLEU metrics \cite{hownot}.

Table~\ref{tab:result2} shows the performance of our model (VED-VAttn-$\bar h$) compared with two main baselines. We see that VEDs are slightly worse than the deterministic encoder-decoder (DED) in this experiment. However, variational attention outperforms deterministic attention in terms of both quality and diversity, showing that our model is effective in different applications. However, we find the improvement is not so large as in the previous experiment. We conjecture that in conversational systems, there is a weaker alignment between the source and target information. Hence, the attention mechanism itself is less effective. 

\section{Related Work}

The variational autoencoder (VAE) was proposed by \newcite{kingma2013auto} for image generation. In NLP, it has been used to generate sentences~\cite{Vseq2seq}. \newcite{VHRED} propose a variational encoder-decoder (VED) model to generate better (more diverse and thus meaningful) replies in a dialog system. VED frameworks have also been applied to knowledge base reasoning \cite{VIKB}. Another thread of VAE/VED applications is to control some characteristics of generated data, such as the angle of a face image~\cite{disentangle}, and the sentiment of a sentence~\cite{control}. 

In this paper, the focus is on the scenario where VED is combined with attention mechanism. We show that the variational attention space is effective, in terms of the diversity of sampled sentences (since VEDs are probabilistic models). Although previous studies have addressed diversity using diversified beam search~\cite{DBS} and determinantal point processes~\cite{dpp}, we would like to point out that our paper is ``orthogonal'' to those studies. The diversity in our approach arises through probabilistic modeling, as opposed to a manually specified heuristic function of the diversity metric. It is to be noted that our approach can be naturally combined with the above methods. 
 
\section{Conclusion and Future Work}
In this paper, we proposed a variational attention mechanism for variational encoder-decoder (VED) frameworks. We observe that, in VEDs, if the decoder has direct access to the encoder, the connection may bypass the variational space. Traditional attention mechanisms might serve as bypassing connection, making the output less diverse. Our variational attention mechanism imposes a probabilistic distribution on the attention vector. We also proposed different priors for the attention vector. The proposed model was evaluated on two tasks: question generation and dialog systems, showing that variational attention yields more diversified samples while retaining high quality.

In future work, it would be interesting to investigate VEDs that model the attention probability with Dirichlet distributions (see Figure~\ref{fig:geo}d). Our framework also provides a principled methodology for designing variational encoding-decoding models without the bypassing phenomenon.

\section*{Acknowledgments}
We thank Hao Zhou for helpful discussions. The Titan Xp GPU used for this research was donated by the NVIDIA Corporation to Olga Vechtomova.

\bibliographystyle{acl}
\bibliography{coling}

\end{document}